\documentclass[runningheads]{llncs}

\usepackage{eccv}

\usepackage{eccvabbrv}

\usepackage{graphicx}
\usepackage{booktabs}
\usepackage{placeins}

\usepackage[accsupp]{axessibility}  % Improves PDF readability for those with disabilities.

\usepackage{hyperref}

\usepackage{orcidlink}

\DeclareMathOperator*{\argmin}{arg\,min}

\begin{document}

% ---------------------------------------------------------------

\title{Conditional Neural Optimal Transport for Predicting Cellular Phenotypes from Molecular Structure}

\titlerunning{Conditional Neural Optimal Transport for Cellular Phenotypes}

\author{
Gauthier Avite\inst{1,*} \and
Maxime Sanchez-Renauld\inst{1,2,3,4,5,*} \and
Nicolas Bourriez\inst{1,*} \and
Auguste Genovesio\inst{1}
}

\authorrunning{G. Avite et al.}

\institute{
IBENS, Ecole Normale Supérieure, Université PSL, Paris, France
\and
Institut Curie, Université PSL, Paris, France
\and
INSERM, Paris, France
\and
Mines ParisTech, Université PSL, Paris, France
\and
Iktos, Paris, France\\
\email{\{gauthier.avite,maxime.sanchez,nicolas.bourriez,auguste.genovesio\}@ens.fr}\\[0.4ex]
$^{*}$These authors contributed equally to this work.
}

\maketitle

\begin{abstract}

High-content microscopy enables systematic profiling of cellular responses to chemical perturbations, but the scale of the chemical space makes exhaustive phenotypic characterization experimentally infeasible. This motivates computational models that can predict image-derived phenotypes without acquiring the corresponding treated cells. We formulate molecule-induced phenotype prediction as an inductive conditional transport problem in image representation space. Given a negative-control phenotype and the structure of a molecule, we aim to predict the phenotype induced by the corresponding molecule. We first evaluate classical optimal transport baselines and show that static couplings do not yield useful predictions on large-scale phenotypic image datasets. We then introduce a molecule-conditioned Neural Optimal Transport (NOT) model with a Monge-Gap regularization training objective that learns to transport negative-control unperturbed phenotypes toward perturbed phenotypes using molecular structure as conditioning information. NOT recovers molecule-specific phenotypic effects while reducing microscopy-associated technical variation, thereby facilitating comparisons across experimental batches. On unseen active molecules, the model outperforms baseline approaches, demonstrating that chemically conditioned transport can generalize beyond the molecules observed during training. We identified the molecular encoder as the main limitation to this generalization, while transport in a compressed representation space improves performance and scalability. These results establish NOT as a promising framework for predicting cellular phenotypes from molecular structure and negative-control phenotypes, while highlighting the development of more informative molecular representations as a key direction for improving out-of-distribution performance.

\end{abstract}

\section{Introduction}
\label{sec:intro}

Cell Painting assay~\cite{bray2016cellpainting} provides rich image-based descriptions of cellular responses to chemical and genetic perturbations by staining multiple cellular compartments. Large-scale initiatives such as JUMP-CP~\cite{chandrasekaran2023jump} now provide standardized morphological profiles for hundreds of thousands of perturbations (see Appendix~\cref{appendix:cell_painting} for additional background on the assay and dataset). However, the chemical space of drug-like molecules remains far larger than any imaging campaign can cover. Acquiring a Cell Painting-based phenotypic response for every candidate molecule is therefore experimentally and economically infeasible, creating a need for models that can predict these image-derived phenotypes without acquiring the corresponding treated cells.

We study this problem in phenotypic representation space. Cell Painting images are encoded into self-supervised DINOv2 representations~\cite{oquab2024dinov2} and aggregated at the well level. Given negative-control DMSO phenotypes and the structure of a molecule, our objective is to predict the corresponding perturbed phenotypes. Our model acts on individual representations but is trained at the distribution level by matching predicted and observed phenotype sets. Because control and treated wells are unpaired, the problem is naturally distributional rather than a pointwise regression task. It is also inductive: a useful model must predict responses for molecules not observed during training.

Existing methods address only parts of this setting. Contrastive image-structure models~\cite{yuan2023cloome,fradkin2024molphenix} align molecules and acquired microscopy profiles in a shared space, but do not directly predict the distributional shift from negative-control to perturbed phenotypes. Generative approaches instead aim to synthesize perturbation-induced images, whereas our model predicts phenotypic representations and is trained by matching predicted and observed phenotype distributions. Classical optimal transport can align observed control and perturbed distributions, but the resulting coupling is transductive and provides no prediction rule for unseen molecules. Structure-to-phenotype prediction is further complicated by activity cliffs, for which small structural changes can induce large phenotypic shifts~\cite{vanTilborg2022activitycliffs,sanchez2026activitycliffs}.

We therefore adapt molecule-conditioned Neural Optimal Transport to predict Cell Painting phenotypes. Building on the Monge-Gap estimator~\cite{uscidda2023monge} and its conditional extension by Driessen et al.~\cite{driessen2026conditional}, we learn an inductive map that transforms individual negative-control phenotypes under molecular conditioning, while matching the resulting predictions to the observed perturbed phenotype distribution. While Driessen et al. apply conditional Monge-Gap transport to single-cell transcriptomic responses, we consider well-level image-derived phenotypes, which are high-dimensional and strongly affected by plate- and batch-level technical variation. We adapt the framework by transporting autoencoded DINOv2 representations, injecting the molecular condition through multi-head attention, and incorporating an unbalanced-OT resampling heuristic~\cite{eyring2024unbalanced} to mitigate over-represented or noisy regions across experimental batches.

We evaluate the resulting model under two complementary forms of generalization. On held-out plates containing molecules observed during training, Neural OT recovers molecule-specific phenotypic effects and generates profiles that are more comparable across experimental batches than either raw or DMSO-normalized measurements. On held-out active molecules, the model outperforms random and identity baselines, providing evidence that molecular conditioned transport can generalize beyond the compounds observed during training. The Monge-Gap regularizer, making the learned map an approximate optimal transport rather than an arbitrary distribution-matching map, governs a trade-off between the two retrieval granularities: it markedly improves replicate-level retrieval, while at plate level a small penalty leaves performance within noise of the best and stronger regularization becomes detrimental. Our ablations further show that transport in a compressed phenotype space via autoencoding improves performance, whereas compressing the molecular condition substantially reduces retrieval, and identify molecular representation quality as the main possible improvement to out-of-distribution generalization.

\begin{figure}
    \centering
    \includegraphics[width=1\linewidth]{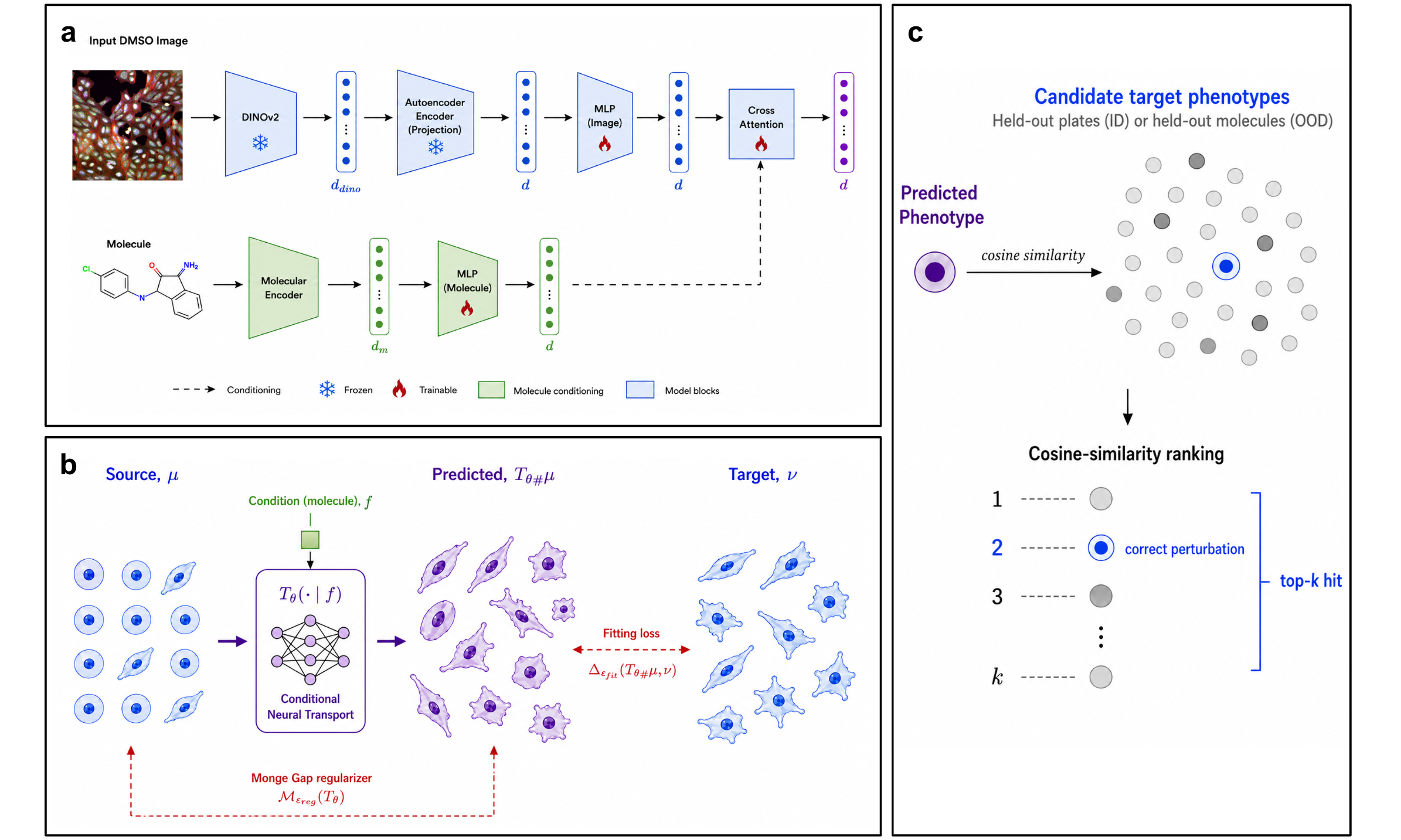}
    \caption{
\textbf{Overview of the molecule-conditioned Neural Optimal Transport framework.}
\textbf{(a)} \textit{Model architecture}. DMSO images and molecular structures are encoded separately and projected to a common dimension \(d\), then combined through cross-attention to predict perturbation-induced phenotype embeddings.
\textbf{(b)} \textit{Training objective}. The conditional map \(T_{\theta}(\cdot \mid f)\) transforms the source distribution \(\mu\) into \(T_{\theta\#}\mu\). Learning minimizes a fitting loss between the predicted and target distributions, together with a Monge-Gap regularizer that constrains the transport.
\textbf{(c)} \textit{Evaluation}. Predictions are ranked against candidate target phenotypes by cosine similarity. Retrieval is successful when the correct perturbation appears among the top \(k\) candidates, using held-out plates for in-distribution evaluation and held-out molecules for out-of-distribution evaluation.
}
    \label{fig:scheme}
\end{figure}

\section{Related Work}
\label{sec:related-work}

\subsection{Contrastive and generative phenotype models}
Contrastive methods such as CLOOME, MoCoP, and MolPhenix \cite{yuan2023cloome,nguyen2023mocop,fradkin2024molphenix} align molecular structures and microscopy phenotypes in a shared space for cross-modal retrieval. They can generalize molecular representations to unseen molecules, but return points in a joint alignment space scored against acquired images rather than phenotype distributions in the native morphology space, and do not model the shift from negative-control morphology. Generative approaches including PhenDiff, CellFlux, IMPA, and MorphoDiff \cite{bourou2024phendiff,zhang2025cellflux,palma2025impa,navidi2025morphodiff} instead simulate or synthesize perturbation-induced morphology. Closer to response prediction, chemCPA \cite{hetzel2022chemcpa} conditions a compositional autoencoder on molecular structure to predict single-cell transcriptomic responses to unseen drugs; this paradigm was subsequently recast as conditional OT \cite{driessen2026conditional}. Our task differs from both cross-modal retrieval and image synthesis: we learn a molecule-conditioned map that transports unpaired control phenotypes toward the distribution of observed perturbation responses in phenotypic-representation space.

\subsection{Optimal transport for cross-domain alignment}
Optimal transport provides a principled framework to compare and align probability distributions \cite{villani2008optimal,peyre2019computational}. Classical Wasserstein distances rely on a ground cost between samples, while Gromov--Wasserstein (GW) compares intra-domain relational structures and can therefore align distributions supported on different metric spaces \cite{memoli2011gromov}. Fused Gromov--Wasserstein (FGW) combines feature-level and structural costs, making it attractive for structured and multimodal alignment when partial feature information is available \cite{vayer2019structured}. These methods are appealing for matching heterogeneous biological representations, but their static formulation is \textit{transductive}: the coupling is computed only for the samples observed during optimization and yields no prediction rule for molecules unseen at optimization time. They can also be computationally expensive, with GW-style solvers relying on non-convex optimization and costly pairwise distance matrices \cite{peyre2016gromov}. Low-rank approximations reduce this cost \cite{scetbon2022linear}, but the resulting alignment still depends strongly on the compatibility of source and target geometries. When some correspondences are known, a line of work injects this partial supervision directly into the transport problem: keypoint-guided OT constrains the plan to preserve a set of annotated matches \cite{gu2022keypoint}, augmented GW adds feature-level priors while relaxing the isometry invariance of GW \cite{demetci2024breaking}, and seed-informed GW fixes known node matchings to guide network alignment \cite{li2023generalized}. We evaluate the FGW instance of this idea, biasing the feature cost at known structure--phenotype pairs, as a static baseline (Appendix~\cref{appendix:static_ot}). All of these formulations nonetheless remain transductive and alignment-oriented, whereas our goal is an inductive map that predicts the perturbed phenotype distribution for unseen molecules.

\subsection{Neural Optimal Transport}
Neural Optimal Transport learns parametric maps that generalize beyond couplings tied to fixed samples. Early approaches parametrize Brenier potentials with input-convex neural networks \cite{brenier1987decomposition,amos2017input} and condition them on perturbations \cite{bunne2022conditional}; CellOT \cite{bunne2023cellot} applies this formulation to unpaired single-cell response prediction. To avoid the convexity constraints and min--max instabilities of these estimators \cite{amos2023amortizing}, the Monge Gap \cite{uscidda2023monge} regularizes an unconstrained map toward optimality, an idea recently extended to conditional single-cell perturbation modeling \cite{driessen2026conditional}. These developments provide the basis for the model detailed in ~\cref{sec:neural-ot}.

\section{Method}
\label{sec:method}

Optimal transport tools underlying our approach are recalled in Appendix~\cref{app:ot-preliminaries}.
Here, we describe how we learn an inductive, molecule-conditioned transport map via the Monge Gap regularizer and an unbalanced resampling heuristic (~\cref{sec:neural-ot}, Figure~\ref{fig:scheme}b), and finally instantiate the model, together with its architecture and conditioning mechanism, for our setting (~\cref{sec:conditional-not}, ~\cref{fig:scheme}a). The classical Gromov--Wasserstein formulations that we use as static-OT baselines are not part of the learned model and are deferred to Appendix~\cref{appendix:static_ot}.

\subsection{Neural Optimal Transport}
\label{sec:neural-ot}

To obtain an \emph{inductive} model, we learn a parametric map $T_\theta$ that can be evaluated on new points rather than a coupling tied to a fixed sample \cite{bunne2023cellot}.

\paragraph{General Neural OT.}
For the quadratic cost $c(x,y)=\tfrac{1}{2}\|x-y\|_2^2$, Brenier's theorem \cite{brenier1987decomposition} states that the optimal Monge map is the gradient of a convex potential, $T^\star=\nabla\phi$ with $\phi$ convex. This motivates parametrizing $\phi$ with an input-convex neural network (ICNN) \cite{amos2017input}, and conditional variants (PICNNs) that condition the potential on side information such as a drug identity \cite{bunne2022conditional}. In practice, enforcing convexity and estimating the associated convex conjugate leads to a min--max objective that is delicate to optimize and can be unstable in high dimension \cite{amos2023amortizing}. Moreover, the Brenier characterization requires the source measure to be absolutely continuous and is tied to the squared-Euclidean cost, which is restrictive for morphology embeddings, where cosine geometry is more natural.

\paragraph{Monge Gap.}
The Monge Gap \cite{uscidda2023monge} removes any architectural constraint on $T_\theta$ and instead regularizes a generic map toward optimality. It relies on the debiased Sinkhorn divergence \cite{feydy2019sinkhorn}
\begin{equation}
    \Delta_{\varepsilon}(\alpha,\beta)
    =
    W_{\varepsilon}(\alpha,\beta)
    -
    \tfrac{1}{2}W_{\varepsilon}(\alpha,\alpha)
    -
    \tfrac{1}{2}W_{\varepsilon}(\beta,\beta),
    \label{eq:sinkhorn-prelim}
\end{equation}
a positive, differentiable divergence that interpolates between OT and Maximum Mean Discrepancy (MMD) and is unbiased for empirical measures. For a source measure $\mu$, the Monge Gap of a map $T_\theta$ is
\begin{equation}
    \mathcal{M}_{\varepsilon}(T_\theta)
    =
    \frac{1}{n}\sum_{i=1}^{n} c\big(x_i,T_\theta(x_i)\big)
    -
    W_{\varepsilon}\!\big(\mu, T_{\theta\#}\mu\big)
    \;\geq\; 0,
    \label{eq:monge-gap-prelim}
\end{equation}
i.e. the excess transport cost of $T_\theta$ relative to the entropic OT optimum between $\mu$ and its own pushforward; it vanishes exactly when $T_\theta$ is $\varepsilon$-optimal for that pair. Learning a Monge map then amounts to minimizing a distributional fit plus this penalty,
\begin{equation}
    \min_\theta\ \Delta_{\varepsilon}\!\big(T_{\theta\#}\mu,\nu\big)
    +
    \lambda\,\mathcal{M}_{\varepsilon}(T_\theta),
    \label{eq:monge-gap-obj-prelim}
\end{equation}
where the first term drives $T_{\theta\#}\mu$ toward $\nu$ and the second encourages a geometry-preserving optimal map rather than an arbitrary matching. This formulation has recently been extended to conditional single-cell perturbation modeling \cite{driessen2026conditional}.

\paragraph{Unbalanced resampling.}
Standard OT assumes that $\mu$ and $\nu$ share the same total mass and matches them exactly, which makes it sensitive to outliers and to over- or under-represented regions, both common across Cell Painting plates and batches. Unbalanced OT relaxes the marginal constraints with soft divergence penalties \cite{chizat2018scaling}. In the neural setting, Eyring et al.\ \cite{eyring2024unbalanced} show that injecting unbalancedness improves unpaired domain translation, and realize it through an OT-based resampling heuristic: an unbalanced coupling $\hat{\pi}_{\tau}$ is estimated on each mini-batch and used to resample source and target points before the balanced map estimator is applied, effectively down-weighting noisy or over-represented samples. We adopt this heuristic, controlled by an unbalancedness parameter $\tau$.

\subsection{Molecule-Conditioned Neural OT}
\label{sec:conditional-not}

We parametrize a molecule-conditioned map $T_\theta(x, f_m)$, where $x$ is a negative-control phenotypic representation and $f_m$ is a molecular representation of the molecule derived from its chemical structure.

\paragraph{Architecture and conditioning mechanism.}

A molecular representation \(h_m\) is first extracted from the structure of molecule \(m\) and projected into the phenotype-conditioning space through a trainable encoder, \[ f_m = g_\psi(h_m). \] The representation \(h_m\) may correspond to a molecular fingerprint, physicochemical descriptors~\cite{rogers2010extended}, or the embedding of a pretrained molecular encoder~\cite{ross2022molformer,ji2024unimol2}. The control phenotype \(x\) and molecular condition \(f_m\) are then combined through a multi-head attention block followed by a feed-forward residual block, producing the transported phenotype \(T_\theta(x \mid f_m)\). We compare alternative molecular representations in Section~\ref{sec:ood}, while keeping the conditional transport architecture fixed. Full architecture and optimization hyperparameters are described in Figure~\ref{fig:scheme}a and reported in Appendix~\ref{app:neural_ot_training}. \\

For a batch of control phenotypes $\{x_i\}_{i=1}^{n}\sim\mu$ and perturbed phenotypes $\{y_j\}_{j=1}^{m}\sim\nu$ associated with that molecule, we minimize:
\begin{equation}
    \mathcal{L}(\theta)
    =
    \Delta_{\varepsilon_{\mathrm{fit}}}
    \left(T_{\theta\#}\mu,\nu\right)
    +
    \lambda
    \left[
    \frac{1}{n}\sum_{i=1}^{n} c\big(x_i,T_\theta(x_i\,|\,f_m)\big)
    -
    W_{\varepsilon_{\mathrm{reg}}}
    \left(\mu,T_{\theta\#}\mu\right)
    \right],
    \label{eq:method-monge-gap}
\end{equation}
which is the Monge-Gap objective~\eqref{eq:monge-gap-obj-prelim} instantiated with the conditioned map, two separate entropic scales $\varepsilon_{\mathrm{fit}}$ and $\varepsilon_{\mathrm{reg}}$, and a cosine ground cost $c$ that matches the geometry of DINOv2 phenotypic representations. When unbalanced resampling is enabled, the batches $\{x_i\}$ and $\{y_j\}$ are replaced by their $\hat{\pi}_{\tau}$-resampled counterparts before Equation~\eqref{eq:method-monge-gap} is evaluated. Both terms are differentiable in $\theta$. At inference, the phenotypic response of an unseen molecule is predicted by pushing its negative-control unperturbed phenotype through $T_\theta(\cdot\,|\,f_m)$.

\section{Data and Evaluation Protocol}

\subsection{Data}
\label{sec:data}

Cell Painting is a high-content fluorescence microscopy assay that captures cellular phenotypes by staining multiple cellular compartments under chemical or genetic perturbations~\cite{bray2016cellpainting}. We use the JUMP Cell Painting dataset (JUMP-CP), a large-scale, standardized collection of perturbation experiments performed across multiple sites and experimental batches in U2OS cells~\cite{chandrasekaran2023jump}. Following~\cite{sanchez2026activitycliffs}, each five-channel microscopy image is converted into a single image-level phenotypic representation using DINOv2~\cite{oquab2024dinov2}. The image-level representations acquired from the same experimental well are then aggregated into one well-level phenotype. Each well therefore constitutes one experimental replicate of the corresponding perturbation.

Each molecule plate contains DMSO negative-control wells and the same eight positive-control molecules. The DMSO profiles define the source phenotype distributions transported by the model. Because the eight positive controls are repeated across plates, laboratories, and experimental batches, they provide a controlled benchmark for evaluating generalization across held-out plates while keeping molecule identity fixed. Examples of Cell Painting images from these controls are shown in Figure~\ref{fig:exemple_img_jump_ctrl}.

\subsection{Evaluation Splits}
\label{sec:splits}
We evaluate two complementary settings that probe different notions of generalization; in both, source samples are drawn from the negative-control wells of the corresponding plates.

\paragraph{Plate-level split (in-distribution).} Training and evaluation plates are disjoint, while the same eight positive-control molecules appear in both sets. All molecules are therefore observed during training, whereas the experimental plates and associated batch effects are held out. This setting measures generalization across plates and batches and provides replicate-based experimental references. It also provides strong replicate-based experimental upper bounds. We use 5-fold cross-validation over plates and train each fold with three random seeds. Each training split contains on average 260,262 DMSO points and 142,196 non-DMSO positive-control points, while each validation split contains 65,065 DMSO points and 35,549 non-DMSO positive-control points.

\paragraph{Molecule-level split (out-of-distribution).} Training and evaluation molecules are disjoint, so at test time the model conditions on the structure of a molecule it has never seen in training. This is the harder and more application-relevant setting, as it directly measures structure-driven generalization to unseen molecules. We use 5-fold cross-validation over molecules and train each fold with 3 random seeds. The benchmark is restricted to the \(12{,}400\) molecules in the top \(10\%\) of phenotypic activity, ranked by their distance to matched DMSO controls. Selection details are given in Appendix~\ref{app:most_active}. Each training split contains on average 267,785 DMSO source points and 265,496 non-DMSO perturbation points, while each validation split contains 263,439 DMSO source points and 66,374 non-DMSO perturbation points. In average, each molecule has around 30 points.

\subsection{Evaluation Metrics}
\label{sec:metrics}

As illustrated in Figure~\ref{fig:scheme}c, evaluation is formulated as a retrieval task: predicted phenotypes are ranked against measured candidate phenotypes using cosine similarity, and retrieval is successful when the candidate associated with the correct molecule appears among the top \(k\).

\paragraph{In-distribution retrieval.} For the plate-level split, predictions are evaluated against the eight positive-control molecules present on each held-out plate. We report Retrieval@1, which measures whether the nearest candidate target phenotype corresponds to the correct molecule.

\paragraph{Out-of-distribution retrieval.} For the molecule-level split, predictions are ranked against all candidate molecules in the held-out fold, and we report Retrieval@10 at two granularities. Replicate-level retrieval (R@10\textsubscript{rep}) ranks individual target wells for each predicted well and is therefore sensitive to replicate-level variability. Plate-level retrieval (R@10\textsubscript{plate}) first averages predicted and target phenotypes within each plate for each molecule before ranking, reducing replicate noise and more directly reflecting the downstream objective of molecule prioritization.

We emphasize plate-level retrieval because it is more stable while better reflecting downstream molecule prioritization.

Whenever measured phenotypes are used as queries for the experimental reproducibility references, the corresponding query replicates are excluded from the retrieval catalogue.

\section{Results}
\label{sec:experiments}

\subsection{In-Distribution Generalization Across Experimental Plates}
\label{sec:in-dist}

We first evaluate generalization across held-out plates using the in-distribution split defined in Section~\ref{sec:splits}.

\paragraph{Experimental reproducibility references.}
To contextualize model performance, we construct two references from observed perturbed replicates. For the held-out-replicate reference, the real replicates of each molecule within a test plate are randomly divided into two disjoint sets: one is treated as the prediction and the other as the retrieval target. This avoids self-matching and estimates the agreement that can be expected between independent experimental replicates. We additionally report a plate-mean replicate reference, obtained by averaging observed target replicates within each plate and molecule. As it directly uses the measured perturbed phenotypes, this reference provides an experimental estimate of the maximum performance possible after suppressing replicate-level noise.

\begin{figure}[t]
   \centering

   \begin{subfigure}{0.48\textwidth}
       \centering
       \includegraphics[width=\linewidth]{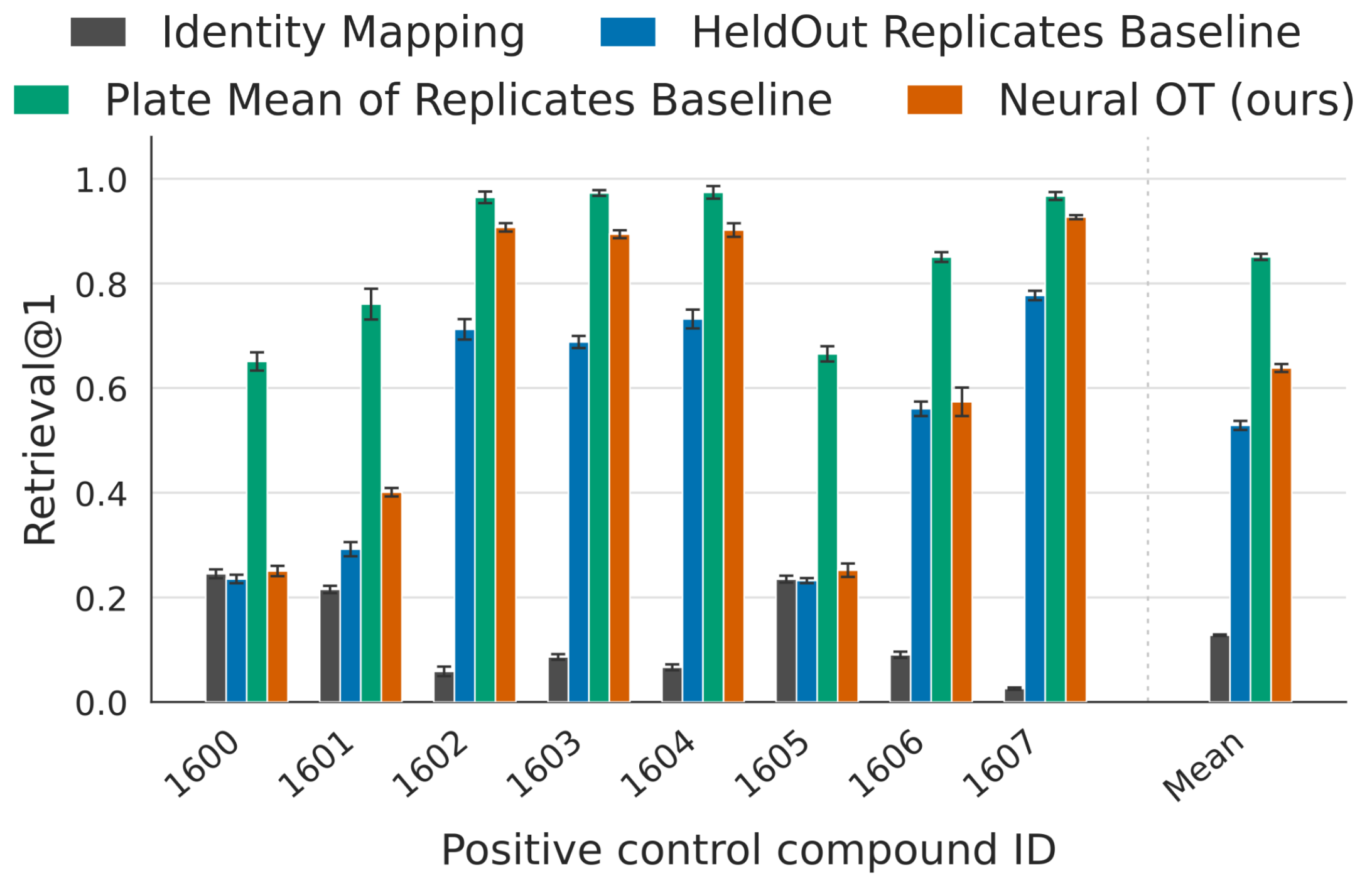}
       \caption{Performance of Neural-OT versus Baselines}
       \label{fig:poscon_ret@1}
   \end{subfigure}
   \hfill
   \begin{subfigure}{0.48\textwidth}
       \centering
       \includegraphics[width=\linewidth]{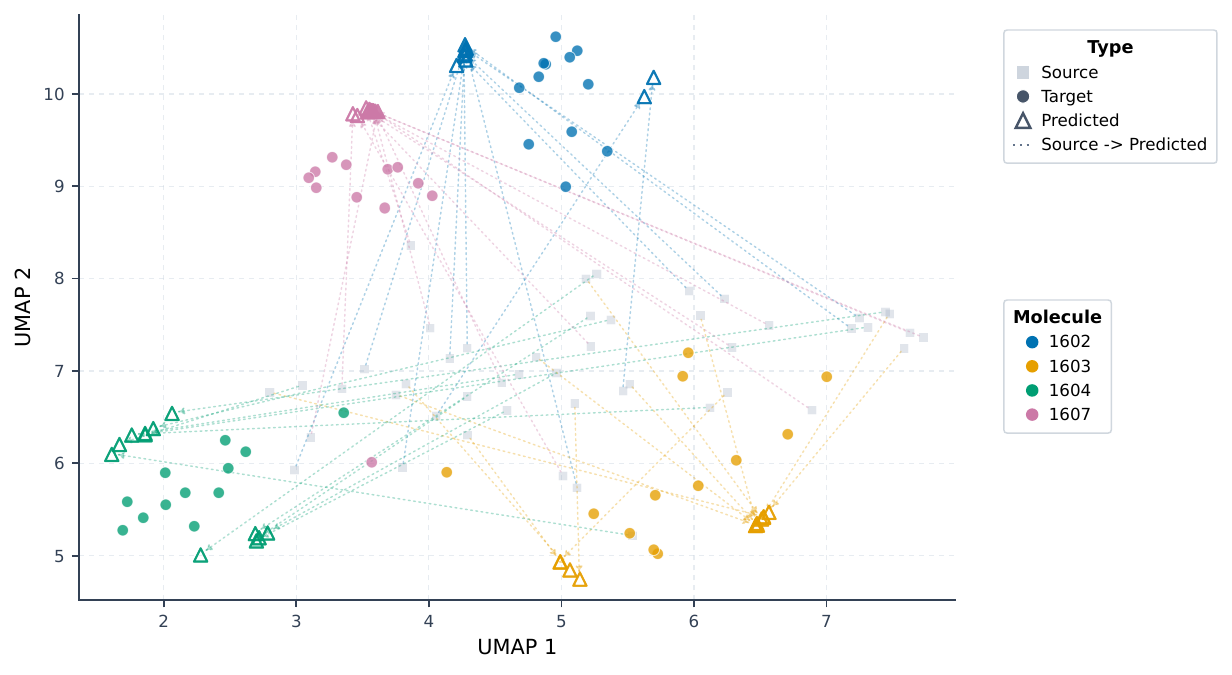}
       \caption{UMAP of Sources, Targets and Predictions}
       \label{fig:umap_ot}
   \end{subfigure}
   \hfill
\caption{\textbf{In-distribution prediction on held-out plates.}
\textbf{(a)} Retrieval@1 for the eight shared positive controls, reported as mean $\pm$ standard deviation across five folds and three seeds.
\textbf{(b)} UMAP of source, target, and predicted phenotypes for the four most active controls. The projection was fitted jointly on source and target embeddings.}
   \label{fig:umap_ctrl_comparison}
\end{figure}

\paragraph{Neural OT recovers molecule-specific phenotypic activities.}
Figure~\ref{fig:poscon_ret@1} compares Neural OT with the identity mapping (no transport) and the two replicate-based references. Leaving the DMSO profiles unchanged results in low Retrieval@1, confirming that the perturbed phenotypes cannot be recovered from the plate-specific controls alone. In contrast, Neural OT improves retrieval for all of the eight molecules and substantially increases the mean Retrieval@1. The improvement is especially pronounced for molecules inducing strong and distinctive morphological responses. Molecules with weaker effects remain harder, as their target distributions overlap more strongly with the DMSO controls and with other perturbations.

For all molecules, Neural OT reaches or exceeds the performance obtained between independently held-out experimental replicates. It nevertheless remains slightly below the plate-mean replicate reference, indicating that part of the remaining error is associated with replicate-level variability. Overall, these results show that the model learns molecule-dependent transport directions that generalize across plates, rather than applying a common transformation to all DMSO controls.

\paragraph{Predicted phenotypes recover molecule-specific target structure.}
The qualitative visualization in Figure~\ref{fig:umap_ot} supports the retrieval results. The UMAP projection is fitted jointly on the source and target embeddings and then applied to the Neural OT predictions. Neural OT transports controls molecules toward the corresponding target regions and recovers the distinct clusters associated with the strongest phenotypic responses. These results indicate that the model learns molecule-specific transformations rather than a common shift toward a generic perturbed phenotype.

\paragraph{Model-generated phenotypes enable batch-robust cross-plate comparisons.}
Beyond predicting phenotypes that have not been experimentally acquired, a model trained across heterogeneous plates can be used to derive standardized surrogate representations of measured perturbations. For each held-out plate, we generated molecule-conditioned phenotypes from its plate-specific DMSO controls and compared the resulting profiles across plates. Because the same transport map is learned jointly across multiple experimental batches, the generated profiles emphasize perturbation-associated variation that is reproducible across the training data while attenuating plate-specific batch-related variations.

\begin{figure}[t]
    \centering
    \includegraphics[width=\textwidth]{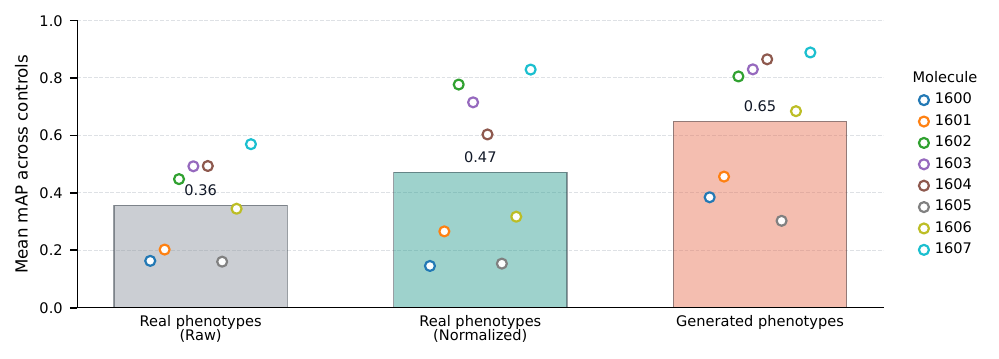}
    \caption{
    \textit{Generated phenotypes improve cross-plate comparability.}
    Cross-plate retrieval performance on held-out plates. Bars show the mean mAP across controls, while points denote individual molecules. Real phenotypes are evaluated either directly or after per-plate DMSO-based normalization. Generated phenotypes achieve higher cross-plate mAP than both real baselines.}
    \label{fig:batch_effect}
\end{figure}

We evaluated this property using cross-plate retrieval, in which each query profile was matched to molecule-level mean profiles from other plates using cosine similarity. Generated phenotypes achieved a mean average precision of \(0.65\), compared with \(0.36\) for raw measured phenotypes and \(0.47\) after DMSO-based per-plate normalization (following standard normalization from \cite{sanchez2026activitycliffs}, Figure~\ref{fig:batch_effect}). These results indicate that model-generated profiles can serve as batch-robust surrogate representations for comparing molecules across experimental batches. 

\subsection{Out-Of-Distribution Regime}
\label{sec:ood}

In the following regime, we focus on predicting the phenotypic response of molecules that were \emph{never seen during training} using the molecule-level split of Section~\ref{sec:splits}.

Our ablation baseline uses \texttt{morganc+rdkc} structure features; NOT with a cosine ground cost; no dimensionality reduction on the conditioning signal; DINOv2 phenotype embeddings reduced to 50D by an autoencoder, the reduction that performs best both in the in-distribution setting (Section~\ref{sec:in-dist}) and out-of-distribution (Table~\ref{tab:dim-reduction}b); Monge-Gap weight $\lambda=0.1$; and unbalancedness $\tau=1$. We compare it against three reference mappings: the \emph{identity} map, which predicts the negative-control morphology unchanged; a \emph{plate-mean} predictor, which assigns every compound its plate-average phenotype; and a \emph{random} baseline. Full ablation hyperparameters are listed in Appendix~\ref{app:ood_training}.

\begin{figure}[t]
    \centering
    \begin{subfigure}[t]{0.49\textwidth}
        \centering
        \includegraphics[width=\linewidth]{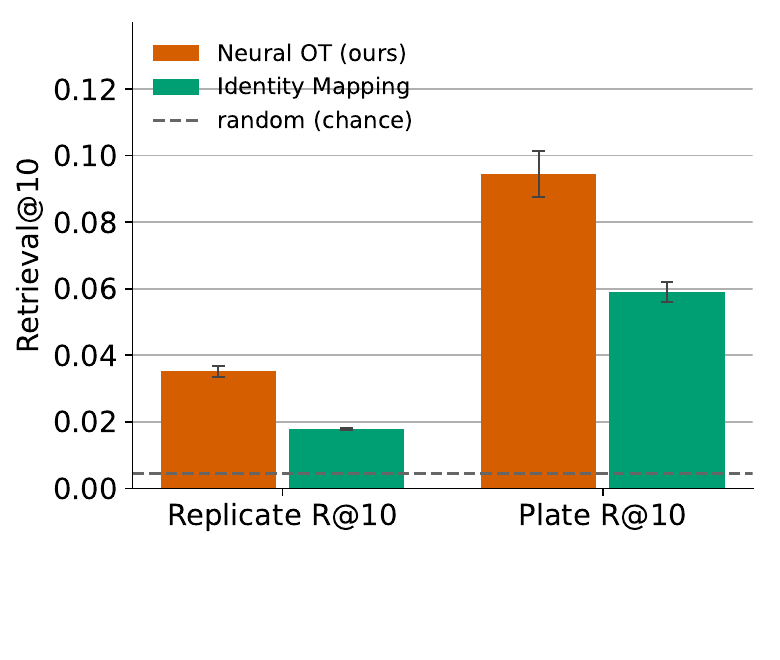}
        \caption{NOT vs.\ baselines}
        \label{fig:ood-baselines}
    \end{subfigure}
    \hfill
    \begin{subfigure}[t]{0.49\textwidth}
        \centering
        \includegraphics[width=\linewidth]{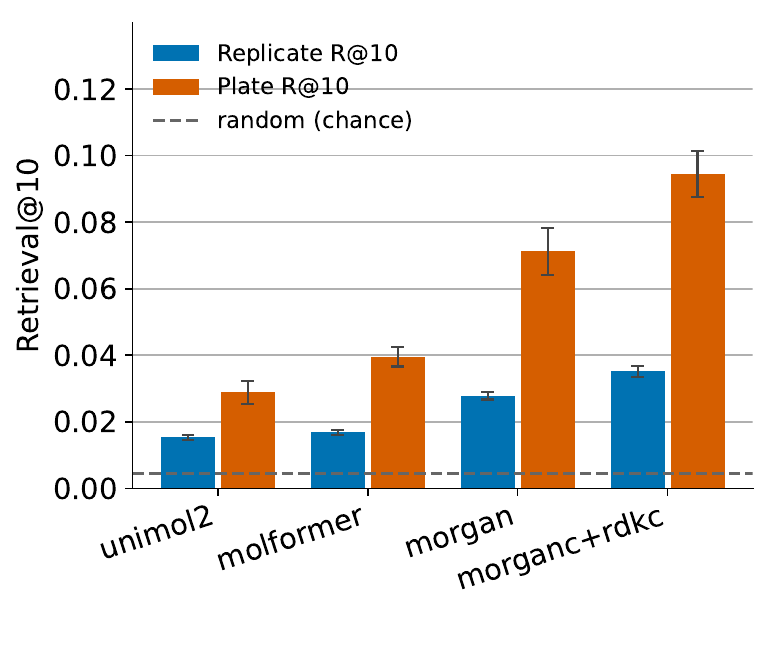}
        \caption{Effect of the structure encoder}
        \label{fig:ood-encoder}
    \end{subfigure}
    \caption{\textit{Neural OT on the molecule-level OOD split (5-fold, 3-seed mean $\pm$ std).} \textbf{(a)} Replicate- and plate-level Retrieval@10 for NOT (ours) and the identity mapping: the conditioned map clearly outperforms the baseline. \textbf{(b)} The same metrics for four molecular structure encoders under an otherwise identical NOT, ordered by plate-level retrieval: retrieval improves substantially with encoder quality. Bars show the mean and error bars the standard deviation over five folds and three seeds. The dashed line marks the chance level ($k/M=10/2248$).}
    \label{fig:ood-main}
\end{figure}

\paragraph{Neural OT generalizes to unseen compounds.}
Figure~\ref{fig:ood-baselines} contrasts NOT with the identity and plate-mean baselines on the ablation configuration. NOT reaches R@10\textsubscript{plate}$=0.095\pm0.007$, well above the identity map ($0.059\pm0.003$) and the plate-mean predictor ($0.004$); the same ordering holds at replicate level ($0.035$ vs.\ $0.018$ vs.\ $0.004$). The plate-mean predictor is a \emph{constant} map, so it returns the same top-$k$ for every query and exactly $k$ of the $M$ validation compounds are retrieved: it sits precisely at the chance level $k/M=10/2248\approx0.004$, by construction and without run-to-run variance (dashed line in Figure~\ref{fig:ood-main}). A conditioned transport map therefore recovers compound-specific phenotypic structure that neither a trivial copy of the control nor a plate summary can capture.

\paragraph{The molecular encoder is the dominant factor.}
Because the compound enters the map only through the structure encoder, its representation quality governs out-of-distribution behavior. Figure~\ref{fig:ood-encoder} compares four encoders under an otherwise identical NOT (the corresponding numbers are given in Table~\ref{tab:ood-encoders-cv}, Appendix~\ref{app:ood_ablations}): retrieval improves markedly from \texttt{unimol2} and \texttt{molformer} to the fingerprint-based \texttt{morgan}, and is best with the combined \texttt{morganc+rdkc} descriptor, with R@10\textsubscript{plate} rising from $0.029$ to $0.095$. This confirms that conditioning quality is the main bottleneck for generalization. We did not evaluate a large pretrained graph encoder in this study; given the trend, a stronger molecular graph representation \cite{sypetkowski2024scalability} is a promising direction for further gains.

\paragraph{Ablation study.}
We now vary one component at a time around the baseline; all numbers are 5-fold, 3-seed means.

\medskip
\noindent\emph{Loss and cost} (Table~\ref{tab:loss-cost-cv}, Appendix~\ref{app:ood_ablations}). Under cross-validation, the cosine and squared-Euclidean ($\ell_2$) ground costs perform comparably, with error bars that overlap at both granularities, so we do not find a decisive advantage for either. We retain the cosine cost for its geometric consistency with the DINOv2 morphology embeddings, on which cosine similarity is the natural notion of proximity.

\medskip
\noindent\emph{Dimensionality reduction} (Table~\ref{tab:dim-reduction}). Compressing the \emph{conditioning} signal sharply hurts retrieval (Table~\ref{tab:dim-reduction}a): plate R@10 falls from $0.095$ to $0.036$--$0.037$ under an autoencoder or PCA, supporting the view that the map needs a detailed representation of the compound. Compressing the \emph{phenotype} embeddings, in contrast, helps (Table~\ref{tab:dim-reduction}b): a 50D autoencoder reaches $0.101$, ahead of PCA ($0.075$) and of no reduction ($0.066$), consistent with performing OT in a lower-dimensional space to mitigate its well-known curse of dimensionality~\cite{peyre2019computational}.

\begin{table}[h]
\centering
\caption{Impact of dimensionality reduction on OOD retrieval (5-fold, 3-seed mean$\pm$std; best in bold, second underlined). \textbf{(a)} Reducing the \emph{conditioning} embedding sharply hurts retrieval. \textbf{(b)} Reducing the \emph{phenotype} embedding helps, the 50D autoencoder being best; this ablation was run at $\tau{=}0.95$, the near-optimal unbalancedness of Table~\ref{tab:unbalanced-cv} (Appendix~\ref{app:ood_ablations}).}
\small
\setlength{\tabcolsep}{2pt}
\begin{subtable}[t]{0.49\textwidth}
\centering
\begin{tabular}{lcc}
\toprule
Reduction & R@10\textsubscript{rep} & R@10\textsubscript{plate} \\
\midrule
None         & \textbf{0.035{\scriptsize$\pm$0.002}} & \textbf{0.095{\scriptsize$\pm$0.007}} \\
Autoencoder  & 0.017{\scriptsize$\pm$0.001} & 0.036{\scriptsize$\pm$0.002} \\
PCA          & \underline{0.018{\scriptsize$\pm$0.001}} & \underline{0.037{\scriptsize$\pm$0.003}} \\
\bottomrule
\end{tabular}
\caption{Conditioning embedding ($\tau{=}1$)}
\label{tab:cond-dim-cv}
\end{subtable}
\hfill
\begin{subtable}[t]{0.49\textwidth}
\centering
\begin{tabular}{lcc}
\toprule
Reduction & R@10\textsubscript{rep} & R@10\textsubscript{plate} \\
\midrule
None         & 0.024{\scriptsize$\pm$0.001} & 0.066{\scriptsize$\pm$0.004} \\
Autoencoder  & \textbf{0.037{\scriptsize$\pm$0.002}} & \textbf{0.101{\scriptsize$\pm$0.007}} \\
PCA          & \underline{0.027{\scriptsize$\pm$0.001}} & \underline{0.075{\scriptsize$\pm$0.006}} \\
\bottomrule
\end{tabular}
\caption{Phenotype embedding ($\tau{=}0.95$)}
\label{tab:emb-dim-cv}
\end{subtable}
\label{tab:dim-reduction}
\end{table}
% \medskip
\noindent\emph{Regularization and conditioning} (Table~\ref{tab:lambda-cond}). The Monge-gap regularizer trades off the two granularities (Table~\ref{tab:lambda-cond}a): increasing $\lambda$ improves replicate retrieval (best at $\lambda=10$) but degrades plate retrieval, which is highest at low regularization ($\lambda\in\{0,0.1\}$); we keep a small $\lambda=0.1$ for a mild geometry-preserving penalty. The conditioning mechanism matters as well (Table~\ref{tab:lambda-cond}b): injecting the molecular structure through multi-head attention outperforms plain concatenation ($0.095$ vs.\ $0.084$ at plate level), supporting our attention-based design. Unbalanced resampling gives only small, consistent plate gains around $\tau\approx0.90$--$0.95$ (Appendix~\ref{app:ood_ablations}, Table~\ref{tab:unbalanced-cv}).

We conjecture that with a larger $\lambda$, predictions for the same molecule may become more similar to each other, while their average may represent the true molecular effect less accurately. This could be verified by looking at the intramolecular dispersion of predictions across different strengths of Monge-Gap regularization.

\begin{table}[!htbp]
\centering
\caption{Ablations on the Monge-Gap strength and the conditioning mechanism (5-fold, 3-seed mean$\pm$std; best in bold, second underlined). \textbf{(a)} Monge-Gap strength $\lambda$: the geometry-preserving penalty trades replicate for plate retrieval. \textbf{(b)} Injecting the molecular structure by multi-head attention versus plain concatenation.}
\small
\setlength{\tabcolsep}{2pt}
\begin{subtable}[t]{0.49\textwidth}
\centering
\begin{tabular}{lcc}
\toprule
$\lambda$ & R@10\textsubscript{rep} & R@10\textsubscript{plate} \\
\midrule
10.0 & \textbf{0.061{\scriptsize$\pm$0.005}} & 0.062{\scriptsize$\pm$0.005} \\
1.0  & \underline{0.050{\scriptsize$\pm$0.004}} & 0.073{\scriptsize$\pm$0.007} \\
0.1  & 0.035{\scriptsize$\pm$0.002} & \underline{0.095{\scriptsize$\pm$0.007}} \\
0.0  & 0.034{\scriptsize$\pm$0.001} & \textbf{0.101{\scriptsize$\pm$0.007}} \\
\bottomrule
\end{tabular}
\caption{Monge-Gap strength $\lambda$}
\label{tab:lambda-abl-cv}
\end{subtable}
\hfill
\begin{subtable}[t]{0.49\textwidth}
\centering
\begin{tabular}{lcc}
\toprule
Conditioning & R@10\textsubscript{rep} & R@10\textsubscript{plate} \\
\midrule
Attention & \textbf{0.035{\scriptsize$\pm$0.002}} & \textbf{0.095{\scriptsize$\pm$0.007}} \\
Concat    & \underline{0.029{\scriptsize$\pm$0.001}} & \underline{0.084{\scriptsize$\pm$0.005}} \\
\bottomrule
\end{tabular}
\caption{Conditioning mechanism}
\label{tab:cond-mech-cv}
\end{subtable}
\label{tab:lambda-cond}
\end{table}

\medskip
\noindent\emph{Sinkhorn parameters.} Finally, retrieval is stable across a broad range of the entropic scales $\epsilon_{\mathrm{fit}}$ and $\epsilon_{\mathrm{reg}}$, indicating that the model is not sensitive to their precise setting.

\section{Conclusion}
\label{sec:conclusion}
We introduced a molecule-conditioned Neural OT model to predict individual Cell Painting phenotypes from plate-specific negative controls and molecular structure through distribution-level training, defining phenotype prediction as an inductive conditional transport problem. On held-out experimental plates, the model recovers molecule-specific phenotypic effects and produces surrogate profiles that are more comparable across batches than raw or normalized real ones. On molecules unseen during training, the model outperforms baselines, providing evidence that structure-conditioned phenotype prediction can generalize beyond the compounds observed during training.

Our ablations clarify the main factors controlling this generalization. Multi-head attention improves conditioning quality. Transport in a DINOv2 compressed phenotype space improves retrieval, whereas compressing the molecular condition substantially degrades performance. Molecular representation quality therefore remains a major bottleneck for OOD prediction. The Monge-Gap strength controls a trade-off between replicate-level and plate-level retrieval: stronger regularization improves the recovery of individual replicates.

\paragraph{Limitations.} Absolute retrieval remains modest, reflecting the intrinsic difficulty of structure--phenotype prediction and the prevalence of activity cliffs \cite{vanTilborg2022activitycliffs,sanchez2026activitycliffs}; our study also leaves the potential strongest pretrained graph encoders \cite{sypetkowski2024scalability}, due to it being closed-source, unevaluated. The split holds out molecule identities rather than molecular scaffolds, and thus does not guarantee generalization to structurally distant chemical series. The static optimal-transport couplings we tested fail on this data (Appendix~\ref{appendix:static_ot}), and the model predicts phenotype \emph{embeddings} rather than images, so it does not by itself yield inspectable morphologies.

\paragraph{Future work.} These limitations point to three directions. First, stronger molecular graph representations \cite{sypetkowski2024scalability} or molecular encoder trained/fine-tuned during Neural-OT training should raise the conditioning ceiling identified in Section~\ref{sec:ood}. Second, the same conditional-transport formulation could extend beyond molecules to genetic perturbations, by replacing the molecular encoder with a gene or protein encoder such as Geneformer \cite{theodoris2023geneformer} or ESM-2 \cite{lin2023esm2}, potentially enabling prediction of knockout or over-expression phenotypes. Lastly, coupling the transported embeddings with a decoder back to image space would turn distributional predictions into inspectable phenotypes, connecting our approach with generative phenotype models.

\section*{Software and Data}
We will release the code to ensure reproducibility. All source data used in this study are publicly available as part of the JUMP Cell Painting Consortium dataset (\texttt{cpg0016}). Data access instructions and documentation are available at \url{https://broadinstitute.github.io/jump_hub/}.
% \footnote{\url{http:/to_create}}

\section*{Use of generative AI.}
Generative AI tools were used to improve the clarity and language of the manuscript, assist with code generation and review, and support figure preparation. All generated or modified text, code and figures were inspected, tested, and validated by the authors. The scientific content, analyses, interpretations, and conclusions remain the sole responsibility of the authors.

% % Acknowledgements should only appear in the accepted version.
% \section*{Acknowledgements}

% This work was performed using HPC resources from GENCI–IDRIS (Grant 2025-AD010316962).

\bibliographystyle{splncs04}
\bibliography{main}

\clearpage
\appendix

\section{Optimal Transport Preliminaries}
\label{app:ot-preliminaries}

Let $(\mathcal{X},d_{\mathcal{X}})$ and $(\mathcal{Y},d_{\mathcal{Y}})$ be two metric spaces equipped with probability measures $\mu$ and $\nu$. In our setting, $\mu$ is the distribution of negative-control (DMSO) phenotypic representations and $\nu$ the distribution of molecule-perturbed phenotypic representations. We briefly recall the tools we build on and refer to \cite{villani2008optimal,santambrogio2015optimal,peyre2019computational} for comprehensive treatments.

\paragraph{Monge and Kantorovich problems.}
Given a ground cost $c(x,y)$, the \emph{Monge} problem \cite{monge1781memoire} seeks a map $T:\mathcal{X}\to\mathcal{Y}$ that pushes $\mu$ onto $\nu$ while minimizing the total displacement cost,
\begin{equation}
    T^\star \in \argmin_{T_{\#}\mu=\nu}\ \int_{\mathcal{X}} c\big(x,T(x)\big)\,d\mu(x),
    \label{eq:monge-prelim}
\end{equation}
where $T_{\#}\mu$ denotes the pushforward of $\mu$ by $T$. This problem is non-convex and may admit no solution, in particular when $\mu$ and $\nu$ are discrete with unequal support sizes. The \emph{Kantorovich} relaxation \cite{kantorovich1942translocation} instead optimizes over couplings $\pi\in\Pi(\mu,\nu)$, the set of joint distributions with marginals $\mu$ and $\nu$,
\begin{equation}
    W_c(\mu,\nu)
    =
    \min_{\pi\in\Pi(\mu,\nu)}
    \int_{\mathcal{X}\times\mathcal{Y}} c(x,y)\,d\pi(x,y),
    \label{eq:kanto-prelim}
\end{equation}
which is a linear program, convex in $\pi$, and always admits a solution. For empirical measures $\mu=\sum_{i=1}^{n} a_i \delta_{x_i}$ and $\nu=\sum_{j=1}^{m} b_j \delta_{y_j}$, Equation~\eqref{eq:kanto-prelim} becomes a finite linear program over coupling matrices $\pi\geq 0$ with $\pi\mathbf{1}=a$ and $\pi^\top\mathbf{1}=b$. The optimal $\pi$ aligns mass between observed samples, but it is \emph{transductive}: it is defined only for the points seen at optimization time and does not yield a prediction rule for unseen molecules.

\section{Cell Painting data}
\label{appendix:cell_painting} 

\subsection{Cell Painting Assay}

Cell Painting \cite{bray2016cellpainting} is a standardized fluorescence microscopy assay that captures diverse visual characteristics of cells at single-cell resolution. Cells are labeled using a fixed set of fluorescent markers, each highlighting a different cellular component (compartments and organelles, such as the nucleus, cytoskeleton, mitochondria, endoplasmic reticulum, and plasma membrane), and imaged across multiple channels to produce rich, multi-channel microscopy images. Images are typically acquired across five to six fluorescence channels, producing multi-channel microscopy images that encode diverse aspects of cell morphology, organization, and subcellular structure. Cell Painting is widely used in large-scale biological and pharmaceutical studies, including drug discovery, genetic perturbation screening, and functional genomics. Its central premise is that perturbations that affect similar biological pathways induce similar cellular phenotypes, which can be detected through changes in cell morphology and organization. As a result, Cell Painting has become a key modality for representation learning, where the goal is to learn feature embeddings that capture biologically meaningful variation across perturbations with or without task-specific supervision. From a machine learning perspective, Cell Painting datasets pose several challenges. They are high-dimensional, multichannel, and exhibit strong sources of variation unrelated to biological signal, such as technical variation (batch effects), imaging artifacts, and cell-cycle heterogeneity. Moreover, biological semantics are indirect: labels often correspond to treatments or genes rather than explicit visual concepts. Consequently, evaluating representation quality is non-trivial and typically relies on downstream proxy tasks, such as perturbation matching or gene–molecule association retrieval. These properties make Cell Painting pertinent for understanding whether representation learning methods can move beyond appearance-driven features and capture higher-level, biologically meaningful abstractions.

\subsection{JUMP-CP} 

\begin{figure}
    \centering
    \includegraphics[width=1\linewidth]{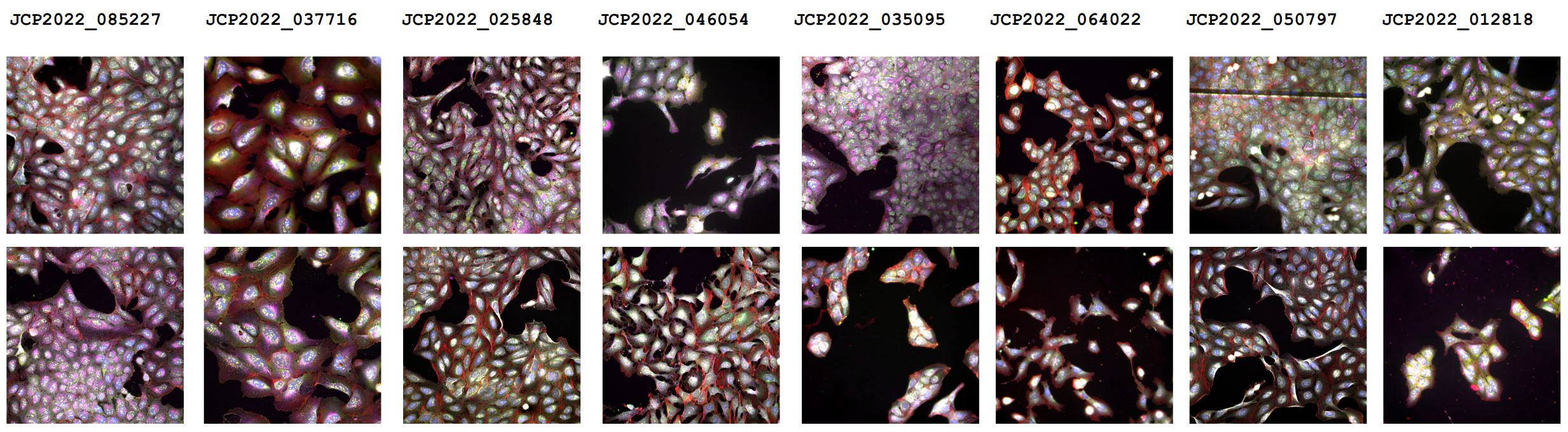}
    \caption{\textbf{Examples of images from the eight positive controls of JUMP-CP.}}
    \label{fig:exemple_img_jump_ctrl}
\end{figure}

JUMP-Cell Painting is a large-scale microscopy dataset generated by the Joint Undertaking for Morphological Profiling (JUMP) Consortium, a collaboration between ten pharmaceutical companies, six technology companies, and two non-profit organizations \cite{chandrasekaran2023jump}. The dataset comprises Cell Painting images of human osteosarcoma (U2OS) cells subjected to diverse perturbations, including chemical treatments, gene overexpression, and CRISPR-Cas9 knockouts. JUMP-CP includes over 116,750 molecules, 12,602 gene overexpression perturbations, and 7,975 gene knockouts, totaling approximately 115 TB of data and capturing single-cell profiles for more than 1.6 billion cells. Each experimental molecule plate, across all batches and laboratories, contains the same eight positive-control molecules (Figure~\ref{fig:exemple_img_jump_ctrl}) and negative controls (DMSO only). We used these shared controls to define a standardized benchmark for evaluating representation robustness across experimental conditions.

\subsection{Selection of the Most Phenotypically Active Molecules}
\label{app:most_active}
\begin{figure}[t]
\centering
\includegraphics[width=\linewidth]{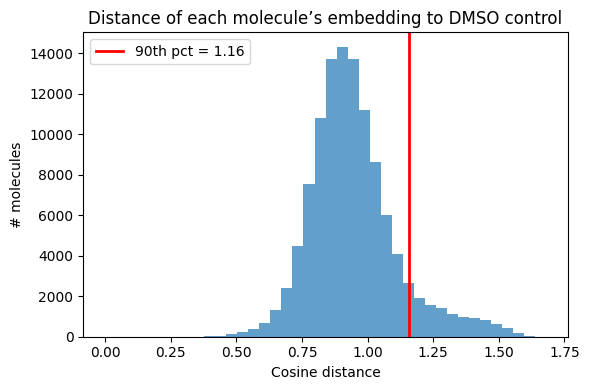}
\caption{\textbf{Distribution of molecule-level phenotypic distances to DMSO negative controls.} The red vertical line indicates the threshold separating the top ($10\%$) most phenotypically active molecules, retained for the out-of-distribution evaluation, from the remaining molecules.}
\label{fig:most_active}
\end{figure}

To focus the out-of-distribution evaluation on compounds that induce a measurable phenotypic response, we retained the top ($10\%$) most phenotypically active molecules. This threshold was motivated by the analysis in \cite{sanchez2026activitycliffs}, which estimated that approximately ($10\%$) of the screened molecules induced a detectable phenotypic response. Molecules were ranked according to the distance between their phenotypic representations and those of the corresponding DMSO negative controls. The resulting activity distribution and the threshold used to define the selected subset are shown in Figure~\ref{fig:most_active}.

The activity score and threshold were computed independently of model predictions and were used exclusively to define the molecules evaluated in the OOD benchmark. No activity score, rank, or measured target phenotype was provided to the model at training or inference time.

\section{Training HP of Neural OT}

\subsection{Implementation Details}
\label{app:implementation}

\paragraph{Phenotypic representations.}
We extract image-level representations using DINOv2-Giant~\cite{oquab2024dinov2} and aggregate the representations belonging to the same experimental well into a single well-level phenotype. Unless otherwise stated, the resulting embeddings are compressed to 50 dimensions using an autoencoder before being provided to the NOT network. We additionally evaluate PCA-based reduction and transport in the unreduced DINOv2 space in Table~\ref{tab:emb-dim-cv}.

\paragraph{Molecular representations.}
We compare four molecular conditioning representations. The first is a binary Morgan fingerprint (ECFP)~\cite{rogers2010extended}. The second concatenates a count-based Morgan fingerprint with RDKit physicochemical descriptors (\texttt{morganc+rdkc}) as done in~\cite{yuan2023cloome}. We additionally evaluate embeddings extracted from two pretrained molecular encoders: MolFormer~\cite{ross2022molformer}, which operates on molecular string representations, and Uni-Mol2~\cite{ji2024unimol2}, which incorporates molecular geometric information. For a controlled comparison, each representation is passed through the same trainable projection network \(g_\psi\), while the remaining transport architecture and optimization procedure are kept unchanged. We specifically used MoLFormer-XL-both-10pct and Uni-Mol2-1.1B.

\subsection{Neural OT Training Details and Hyperparameters}
\label{app:neural_ot_training}

The hyperparameters reported in this section correspond to the \emph{in-distribution} experiments of Section~\ref{sec:in-dist} (the positive-control benchmark); the out-of-distribution ablation of Section~\ref{sec:ood} uses the separate configuration of Appendix~\ref{app:ood_training}. We evaluated our NOT with 5-fold cross-validation over plates, training each fold with three random seeds.
In each fold, 80\% of the plates are used for training and the held-out 20\% for validation, i.e.\ about 1285 training and 322 validation plates per fold.
Image representations were extracted from DINOv2-Giant \cite{oquab2024dinov2} embeddings and compressed to 50 dimensions using an autoencoder. 
The molecular condition was represented by a Morgan fingerprint with radius 2 (ECFP4 \cite{rogers2010extended}) and 1024 bits.
All experiments were performed at the well-embedding level representations of phenotypes.

\begin{table}[t]
    \centering
    \small
    \caption{Experimental setup for NOT training.}
    \label{}
    \begin{tabular}{@{}p{0.42\linewidth}p{0.58\linewidth}@{}}
        \toprule
        Element & Value \\
        \midrule
        Seeds & 1, 2, 3 \\
        Split strategy & Plate-level 5-fold cross-validation \\
        Train/validation split & 80/20 per fold \\
        Number of train/validation plates & $\approx$1285 train, $\approx$322 validation per fold \\
        Image embedding reduction & Autoencoder, 50 dimensions \\
        Molecular condition & Morgan fingerprint \\
        Fingerprint size & 1024 bits \\
        Morgan radius & 2 \\
        Representation level & Well-level embeddings \\
        \bottomrule
    \end{tabular}
\end{table}

The NOT model uses an attention-based conditional architecture. 
The architecture includes both attention and feed-forward residual connections. 

\begin{table}[t]
    \centering
    \small
    \caption{Neural OT model architecture.}
    \label{tab:neural_ot_architecture}
    \begin{tabular}{@{}p{0.42\linewidth}p{0.58\linewidth}@{}}
        \toprule
        Element & Value \\
        \midrule
        Architecture & Attention conditional network \\
        Phenotypic input & AE-reduced DINOv2 embedding, 50 dimensions \\
        Conditional input & Morgan fingerprint, 1024 bits \\
        Fingerprint encoder & MLP: $1024 \rightarrow 512 \rightarrow 512 \rightarrow 50$, ReLU activations \\
        Number of attention heads & 2 \\
        Feed-forward hidden dimension & 64 \\
        EMA decay & 0.99 \\
        \bottomrule
    \end{tabular}
\end{table}

The model was trained for 50 epochs using Adam with an initial learning rate of $10^{-4}$. 
The learning-rate schedule consisted of a warmup phase, a constant phase, and a cosine decay phase. 
Training batches were grouped by perturbation and plate, with a dataloader batch size of one grouped item (molecule). 

\begin{table}[t]
    \centering
    \small
    \caption{Optimization hyperparameters.}
    \label{tab:neural_ot_optimization}
    \begin{tabular}{@{}p{0.38\linewidth}p{0.52\linewidth}@{}}
        \toprule
        Hyperparameter & Value \\
        \midrule
        Number of epochs & 50 \\
        Optimizer & Adam \\
        Learning rate & $10^{-4}$ \\
        Learning-rate scheduler & Warmup + constant + cosine decay \\
        Warmup fraction & 0.2 \\
        Warmup epochs & 10 \\
        Constant LR fraction & 0.3 \\
        Constant epochs & 15 \\
        Cosine decay fraction & 0.5 \\
        Cosine decay epochs & 25 \\
        Noise augmentation & None \\
        \bottomrule
    \end{tabular}
\end{table}

\begin{table}[t]
    \centering
    \small
    \caption{Optimal transport loss configuration.}
    \label{tab:neural_ot_loss}
    \begin{tabular}{@{}p{0.38\linewidth}p{0.52\linewidth}@{}}
        \toprule
        Element & Value \\
        \midrule
        Training objective & Sinkhorn fitting loss + Monge-gap regularization \\
        Cost type & Cosine cost \\
        $\varepsilon_{\mathrm{fit}}$ & 0.01 \\
        Sinkhorn fitting blur & $\sqrt{0.01}=0.1$ \\
        Fitting debias & True \\
        $\varepsilon_{\mathrm{reg}}$ & 0.1 \\
        Monge-gap/Sinkhorn cost blur & $\sqrt{0.1}\approx 0.316$ \\
        Monge-gap debias & False \\
        Regularization strength & 1 \\
        Reach / $\tau$ resampling & None \\
        \bottomrule
    \end{tabular}
\end{table}

\subsection{Out-of-Distribution Ablation Setup and Hyperparameters}
\label{app:ood_training}

The molecule-level out-of-distribution study of Section~\ref{sec:ood} uses a separate sweep, summarized in Table~\ref{tab:ood_setup}. It differs from the in-distribution training above mainly in its data split: molecules, rather than plates, are held out, and performance is estimated by 5-fold cross-validation over 3 seeds. All numbers in Section~\ref{sec:ood} are means $\pm$ standard deviation over the resulting fifteen runs; unless a component is explicitly ablated, it takes the baseline value listed below.

\begin{table}[t]
    \centering
    \small
    \caption{Hyperparameters for the molecule-level OOD ablation (Section~\ref{sec:ood}). The settings that differ from the in-distribution setup of Appendix~\ref{app:neural_ot_training} are the data split, the entropic scales $\varepsilon_{\mathrm{fit}},\varepsilon_{\mathrm{reg}}$, the Monge-gap weight $\lambda$, the number of epochs, and the structure encoder.}
    \label{tab:ood_setup}
    \begin{tabular}{@{}p{0.42\linewidth}p{0.58\linewidth}@{}}
        \toprule
        Element & Value \\
        \midrule
        Split strategy & Molecule-level OOD (disjoint train/test molecules) \\
        Cross-validation & 5-fold $\times$ 3 seeds \\
        molecule subset & Most-active molecules \\
        Phenotype embedding & DINOv2-Giant, autoencoder-reduced to 50D \\
        Structure encoder (baseline) & \texttt{morganc+rdkc}, radius 3; no reduction \\
        Architecture & Attention conditional network; EMA decay 0.99 \\
        Optimizer & Adam \\
        Learning rate & $10^{-4}$, warmup(20\%)+constant(30\%)+cosine(50\%) \\
        Number of epochs & 20 \\
        Objective / cost & Sinkhorn fit + Monge-gap; cosine cost \\
        $\varepsilon_{\mathrm{fit}}$ / $\varepsilon_{\mathrm{reg}}$ & $0.002$ / $0.08$ \\
        Monge-gap $\lambda$ (baseline) & $0.1$ \\
        Unbalancedness $\tau$ (baseline) & $1$ (balanced) \\
        \bottomrule
    \end{tabular}
\end{table}

\subsection{Additional Out-of-Distribution Ablation Tables}
\label{app:ood_ablations}

This section reports, in tabular form, three ablations discussed in Section~\ref{sec:ood}: the structure
encoder (Table~\ref{tab:ood-encoders-cv}, the numbers behind Figure~\ref{fig:ood-encoder}), the ground
cost (Table~\ref{tab:loss-cost-cv}), and the unbalancedness $\tau$ of the resampling heuristic
(Table~\ref{tab:unbalanced-cv}). As everywhere in Section~\ref{sec:ood}, values are means $\pm$ standard
deviation over 5 folds and 3 seeds, using only finished runs.

\begin{table}[h]
\centering
\caption{Structure encoders under NOT on the molecule-level OOD split (5-fold, 3-seed mean$\pm$std). Best in bold, second best underlined. These are the values plotted in Figure~\ref{fig:ood-encoder}.}
\small
\begin{tabular}{lcc}
\toprule
Structure encoder & R@10\textsubscript{rep} & R@10\textsubscript{plate} \\
\midrule
\texttt{morganc+rdkc} & \textbf{0.035{\scriptsize$\pm$0.002}} & \textbf{0.095{\scriptsize$\pm$0.007}} \\
\texttt{morgan}       & \underline{0.028{\scriptsize$\pm$0.001}} & \underline{0.071{\scriptsize$\pm$0.007}} \\
\texttt{molformer}    & 0.017{\scriptsize$\pm$0.001} & 0.040{\scriptsize$\pm$0.003} \\
\texttt{unimol2}      & 0.015{\scriptsize$\pm$0.001} & 0.029{\scriptsize$\pm$0.003} \\
\bottomrule
\end{tabular}
\label{tab:ood-encoders-cv}
\end{table}

\begin{table}[h]
\centering
\caption{Ablation on the training objective and ground cost (5-fold, 3-seed mean$\pm$std). Best in bold, second best underlined.}
\small
\begin{tabular}{lcc}
\toprule
Objective / cost & R@10\textsubscript{rep} & R@10\textsubscript{plate} \\
\midrule
NOT (cosine cost)   & \underline{0.035{\scriptsize$\pm$0.002}} & \underline{0.095{\scriptsize$\pm$0.007}} \\
NOT ($\ell_2$ cost) & \textbf{0.038{\scriptsize$\pm$0.002}} & \textbf{0.095{\scriptsize$\pm$0.006}} \\
\bottomrule
\end{tabular}
\label{tab:loss-cost-cv}
\end{table}

\begin{table}[h]
\centering
\caption{Unbalancedness $\tau$ of the resampling heuristic (5-fold, 3-seed mean$\pm$std). Best in bold, second best underlined. Mild unbalancedness ($\tau\approx0.90$--$0.95$) gives small, consistent plate gains over the balanced $\tau{=}1$ baseline.}
\small
\begin{tabular}{lcc}
\toprule
$\tau$ (unbalanced resampler) & R@10\textsubscript{rep} & R@10\textsubscript{plate} \\
\midrule
1.00 & 0.035{\scriptsize$\pm$0.002} & 0.095{\scriptsize$\pm$0.007} \\
0.95 & \textbf{0.037{\scriptsize$\pm$0.002}} & \underline{0.101{\scriptsize$\pm$0.007}} \\
0.90 & \underline{0.037{\scriptsize$\pm$0.001}} & \textbf{0.101{\scriptsize$\pm$0.007}} \\
\bottomrule
\end{tabular}
\label{tab:unbalanced-cv}
\end{table}

\section{Static Optimal Transport}
\label{appendix:static_ot}

We contrast our learned map against classical static optimal transport couplings, which we use as baselines. Throughout, $\mu$ and $\nu$ denote the negative control and perturbed morphology distributions, $\Pi(\mu,\nu)$ the set of couplings with these marginals (as in Equation~\eqref{eq:kanto-prelim}), and $d_{\mathcal{X}},d_{\mathcal{Y}}$ intra-domain distances. Unlike the neural map of Section~\ref{sec:neural-ot}, these formulations are \emph{transductive}: they return a coupling over the samples observed at optimization time and provide no prediction rule for unseen molecules.

\subsection{GW}
When source and target live in different spaces, so that a pointwise cost $c(x,y)$ is unavailable, the Gromov--Wasserstein (GW) distance \cite{memoli2011gromov} compares \emph{intra-domain} dissimilarities rather than samples directly,
\begin{equation}
    \mathrm{GW}^p(\mu,\nu)
    =
    \min_{\pi\in\Pi(\mu,\nu)}
    \iint \big|d_{\mathcal{X}}(x,x') - d_{\mathcal{Y}}(y,y')\big|^p \, d\pi(x,y)\,d\pi(x',y').
    \label{eq:gw-prelim}
\end{equation}
GW is invariant to isometries of each space, which makes it appealing for aligning heterogeneous biological modalities, but the resulting problem is a non-convex quadratic assignment.

\subsection{FGW}
Fused Gromov--Wasserstein (FGW) \cite{vayer2019structured} interpolates between a feature-level Wasserstein cost and the structural GW term through a trade-off parameter $\alpha\in[0,1]$,
\begin{equation}
    \mathrm{FGW}^p_\alpha(\mu,\nu)
    =
    \min_{\pi\in\Pi(\mu,\nu)}
    (1-\alpha)\!\int c(x,y)\,d\pi
    \;+\;
    \alpha\!\iint \big|d_{\mathcal{X}}(x,x') - d_{\mathcal{Y}}(y,y')\big|^p\, d\pi\,d\pi,
    \label{eq:fgw-prelim}
\end{equation}
where the couplings inside each integral are as in Equations~\eqref{eq:kanto-prelim} and~\eqref{eq:gw-prelim}. FGW suits structured objects for which both feature and relational information matter, and partial supervision can be injected by biasing the feature cost $c$ at known correspondences. GW and FGW are typically optimized by conditional-gradient schemes that solve an entropic OT problem at each iteration, giving an overall $\mathcal{O}(n^3)$ cost \cite{peyre2016gromov}; low-rank couplings and costs can bring this closer to linear in the number of samples \cite{scetbon2022linear}.

\subsection{Semi-supervised FGW}
\label{appendix:semisup_fgw}
In JUMP-CP a subset of structure--phenotype correspondences is actually known: for the imaged molecules we know which molecule was applied to which well. FGW offers a simple way to exploit such partial supervision by biasing its feature cost at the known matches, an instance of the broader idea of injecting known correspondences into (Gromov--)Wasserstein transport \cite{gu2022keypoint,demetci2024breaking,li2023generalized}. Given a set of supervision pairs $\mathcal{S}=\{(i^\star,j^\star)\}$ that link structure sample $i^\star$ to phenotype sample $j^\star$, we build the feature-cost matrix
\begin{equation}
    C_{ij} =
    \begin{cases}
        -\lambda, & (i,j)\in\mathcal{S},\\
        0, & \text{otherwise},
    \end{cases}
    \qquad \lambda>0,
    \label{eq:semisup-cost}
\end{equation}
and solve the fused problem~\eqref{eq:fgw-prelim} with this cost, i.e.\ $T=\mathrm{FGW}_\alpha(C,d_{\mathcal{X}},d_{\mathcal{Y}})$. The negative entries reward the coupling for placing mass on the known correspondences, so that supervision propagates to the unlabeled pairs through the structural GW term; $\lambda$ controls the strength of the prior and $\alpha$ its balance against the geometry. Setting $\mathcal{S}=\varnothing$ recovers the unsupervised FGW baseline.

We evaluated this heuristic as a static baseline for structure$\to$phenotype matching. On JUMP-CP, however, even a large fraction of supervised correspondences yields only marginal gains and the coupling stays close to chance, consistent with a strong manifold mismatch between fingerprint and morphology geometries. Like GW and FGW, the heuristic is moreover \emph{transductive}: the recovered $T$ matches only the observed samples and provides no map for unseen molecules, which is one of the considerations that motivated the inductive Neural OT model of Section~\ref{sec:neural-ot}.

\subsection{Empirical results on JUMP-CP}
\label{appendix:static_ot_results}
We evaluated these static couplings on the most-active subset of JUMP-CP (the top ${\sim}1\%$, i.e.\ $1125$ molecules), matching molecular fingerprints to morphology embeddings. Table~\ref{tab:static_gw_jump} reports the number of correctly recovered structure$\to$phenotype matches. Plain GW recovers essentially no correct match, on par with a random coupling, and injecting supervision up to $80\%$ of the pairs moves the count only from $1$ to $3$ out of $1125$.

This behavior is not a shortcoming of semi-supervised OT in general: injecting known correspondences into GW/OT is known to sharpen alignment when the two domains share a recoverable geometry, whether across single-cell multi-omics \cite{demetci2024breaking}, heterogeneous domains \cite{gu2022keypoint}, or partially seeded networks \cite{li2023generalized}. The near-chance coupling we observe on JUMP-CP is therefore consistent with a strong manifold mismatch between fingerprint and morphology geometries: on Cell Painting, small structural changes can induce large phenotypic shifts (activity cliffs) \cite{vanTilborg2022activitycliffs,sanchez2026activitycliffs}, so the intra-domain distances that GW aligns are not related by an isometry. A structural coupling, even a partially supervised one, then cannot localize the correct phenotype of a molecule. This failure, together with the absence of any out-of-sample prediction rule, is what motivated the inductive NOT model of Section~\ref{sec:neural-ot}.

\begin{table}[t]
\centering
\small
\caption{Static GW/FGW on the most-active JUMP-CP subset ($1125$ molecules): number of correct structure$\to$phenotype matches. Even $80\%$ of supervised correspondences leaves the coupling at chance level.}
\label{tab:static_gw_jump}
\begin{tabular}{@{}lcc@{}}
\toprule
Method & Phenotype distance & \# correct (out of 1125) \\
\midrule
GW & $\ell_2$ & 0 \\
GW & cosine & 1 \\
Random coupling & --- & 1 \\
Semi-supervised FGW ($80\%$ labels, $\lambda{=}100$) & cosine & \textbf{3} \\
\bottomrule
\end{tabular}
\end{table}

\end{document}